# GEMA: An open-source Python library for self-organizing-maps

Álvaro José García Tejedor and Alberto Nogales

CEIEC Research Institure, Universidad Francisco de Vitoria, Ctra. M-515 Pozuelo-Majadahona km 1,800, 28223 Pozuelo de Alarcón, Spain
a.gtejedor@ceiec.es, alberto.nogales@ceiec.es

**Abstract.** Organizations have realized the importance of data analysis and its benefits. This in combination with Machine Learning algorithms has allowed to solve problems more easily, making these processes less time-consuming. Neural networks are the Machine Learning technique that is recently obtaining very good best results. This paper describes an open-source Python library called GEMA developed to work with a type of neural network model called Self-Organizing-Maps. GEMA is freely available under GNU General Public License at GitHub (https://github.com/ufvceiec/GEMA). The library has been evaluated in different a particular use case obtaining accurate results.

**Keywords:** Machine learning, Neural networks, Self-organizing maps.

## 1. Introduction

The increasing availability of big amounts of data and the drop of computational capacity costs allow many hard problems to be solved by applying machine learning (ML) techniques, [1]. Thus, there is a growing need for ML libraries (implementation of algorithms and models), especially, in the open domain. In this scenario, artificial neural networks (ANN), a subset of biologically inspired ML techniques, are becoming more popular among the artificial intelligence community. This is a consequence of their capability to solve different problems and its good performance.

There are many ANN models, but self-organizing maps (SOMs) are fundamentally different in terms of architecture and learning algorithms. SOMs, also known as Kohonen maps, are based on biological studies of the cerebral cortex and were introduced in 1982 by [2], [3]. This model is an ANN with an unsupervised training algorithm that performs non-linear mapping between high dimensional patterns and a discrete bidimensional representation, called feature map, without external guidelines. It is for that reason that SOM has been widely used as a method for pattern recognition, dimensionality reduction, data visualization, and cluster analysis (categorization), [4].

GEMA, which stands for GEnerador de Mapas Autoasociativos (Self-associative Maps Generator in Spanish) is an implementation of Kohonen's maps that builds a SOM from scratch in a two-step process: training and mapping/classification. The training process finds a coherent clustering (a feature map) using a set of input examples by defining and fine-tuning SOM parameters. The mapping process automatically classifies new input data using the trained network from the previous step. Also, GEMA implements facilities to analyze the results with reports and interactive visualization.

## 2. Theoretical methods

SOM performs a mapping from a higher dimensional input space to a lower-dimensional map space through a two-layered fully connected architecture. The input layer is a linear array with as many neurons (elementary components of ANN) as the dimension of the input data vector ($n$). The output layer (or Kohonen layer) consists of a set of neurons, each of them having an associated weight vector of the same dimension as the input data ($n$) and a position in a rectangular grid of arbitrary size ($k$). All weights are arranged in a $n*k*k$ matrix called weight matrix. Fig. 1 shows a typical architecture of a Kohonen map.

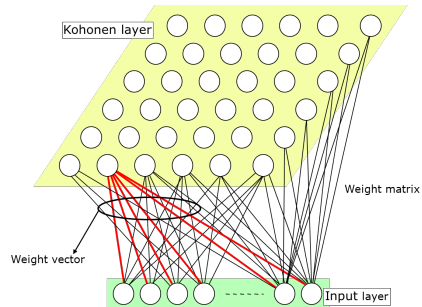
Fig. 1: Kohonen map architecture.

Self-organization is a process described as it follows. A vector from the data space ($X$) is presented to the network. The node with the closest weight vector $W_j$ is the winner neuron or best matching unit (BMU). This is calculated using a simple discriminant function (Euclidean distance) and a "winner-takes-all" mechanism (competition). Then, the unsupervised training algorithm modifies the winner's weight vector depending on its resemblance with the input vector. Input vectors presentation and BMU learning continue until a given number of presentations ($P$) is reached. The result of this iterative process is a trained (self-organized) Kohonen map, represented by a given weight matrix. Each node in the Kohonen layer will answer for a certain pattern previously learned and will recognize all elements belonging to that class. The self-organizing training process guarantee that topological properties of the input space are preserved, and neighbor nodes recognize patterns that share similar characteristics.

## 3. Software Framework

GEMA is a library that has been developed to facilitate the management of Kohonen maps. It allows data scientists to define and train SOMs, using them later to classify new instances from a target dataset. GEMA also helps to analyze the classifier itself and the classification results obtained by visualizing data and obtaining some metrics.

GEMA is written in Python 3.7 with dependencies to some libraries. NumPy[2], a package for scientific computing. Pandas, [5], which is used to manage data structures. Matplotlib described in [6], imageio[3], and Plotly[4] for visualizing the results. Scikit-learn, [7], and SciPy[5] provide more complex mathematical functions. Finally, numba is a Python compiler that accelerates developed functions, [8].

Neural networks require prior dataset manipulation to make it understandable by the network. These processes are grouped into a set of operations called preprocessing, mainly data normalization, although all statistical analysis of the dataset can also be carried out.

The training/learning process involves an incremental adaptation of neurons' weight vectors using a training dataset of unlabeled input vectors until a coherent clustering (a map) is obtained accordingly. A clean SOM is obtained as an instance of class GEMA by a call that sets the map side. It is very usual that once a good SOM has been obtained, the user is interested in saving it and use it in the future. To accomplish this task, the library provides the possibility of saving the information of the map as a JSON. Thus, a pre-trained model can also be loaded with this saved model.

The process to classify a data set with a network using this library is practically the same as the training, except that no weight is modified, and other parameters are not necessary. Only the winning neuron is calculated for each sample to be categorized. The trained map receives unlabeled patterns to be clustered in the space by calculating a discriminant function (for example the Euclidean distance) between each element to be classified and the SOM weight matrix.

---

[2] http://www.numpy.org/

[3] https://github.com/imageio/imageio

[4] https://plotly.com/

[5] http://www.scipy.org/

Finally, a visualization/reporting stage where the user can ask for different plots and reports that provide a friendly interpretation of the results (input dataset and codebooks), cluster analysis and quality measurements. Fig. 2 describes all the methods and the different architectural elements involved.

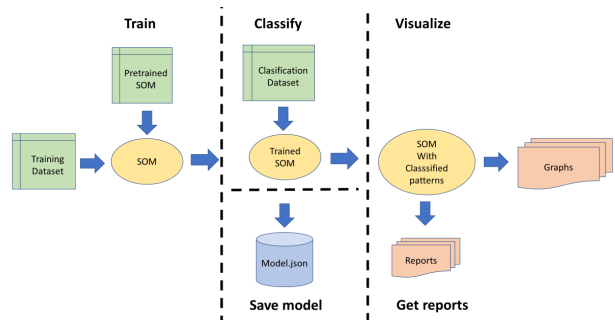

Fig 2. GEMA workflow and interaction between modules.

Fig. 3 shows the same process by implementing the code. In this case a Kohonen map of size 10 is initialized and trained with data from a csv file. The training stage consists of 50000 epochs and uses a learning rate of 0.1. Then, this SOM is used to classify some data instances. At the end, two type of plots have been obtained: a 3D and a 2D heatmap.

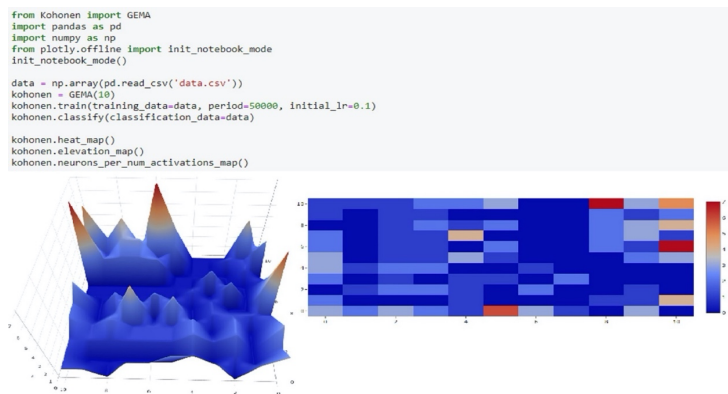

Fig 3. GEMA coding example of a complete workflow.

## 3. Comparison with other tools

Other Python libraries implement Kohonen maps. Kohonen[6] contains some implementations of Kohonen-style vector quantizers although it also supports Neural Gas and Growing Neural Gas. A very simple implementation of a Kohonen map library called som[7]. Somoclu, [9], also works with SOMs but it allows to parallelize the different tasks. A package called PyMVPA for statistical learning analysis includes a class to model SOMs, [10]. NeuPy[8] is a Neural Network library including also a class for Kohonen maps. Another library only for SOMs is SOMPy[9] which follows the structure of the Matlab somtoolbox. MiniSom[10] is a minimalistic implementation of the Self Organizing Maps. Finally, SimpSOM[11] is a lightweight implementation of Kohonen maps.

These libraries implement a lot of the functions provided by GEMA. But in contrast, GEMA has new metrics like topology which provides quality about the density of the map. Users also request for more completed reports containing detailed information of the feature map. Finally, GEMA is the only one providing interactive visualization with graphs like a 3D elevation heatmap or a diagram bar showing how many neurons have been activated a certain number of times.

---

[6] https://github.com/lmjohns3/kohonen
[7] https://github.com/alexarnimueller/som
[8] https://github.com/itdxer/neupy
[9] https://github.com/sevamoo/SOMPY
[10] https://github.com/JustGlowing/minisom
[11] https://github.com/fcomitani/SimpSOM

## 4. Empirical results and evaluation

To test the library, it has been evaluated in three well-known use cases in the field. The first one consists of classifying a set of colours. The second use case consists of classifying images of handwritten digits in black and white. The third use case is formed by three classes of iris plants where each pattern has the length and width of the sepal and the petal. Apart from that, a successful research has been conducted in the field of psychology by classifying students' profiles regarding their competencies, [11]. In the following paragraphs, the experiments for each use case will be explained. The results obtained will be represented in graphs. In the results, it can be seen that patterns with the same characteristics are clustered in similar zones. It should be highlighted that these graphs are not part of the library and have been developed itself only to evaluate the results obtained with the library.

The first use case consists of classifying a set of colors. The patterns for each color are formed by three values between 0 and 255 corresponding to the red, green, and blue channels. A training dataset has been created by generating 500 patterns randomly. Then, a SOM of size 100 has been created and trained. After finishing that stage, a set of 10,000 colors has been classified. The results can be seen in Fig 4. showing that patterns of the same color are distributed in similar zones and the soft transitions of tones among them.

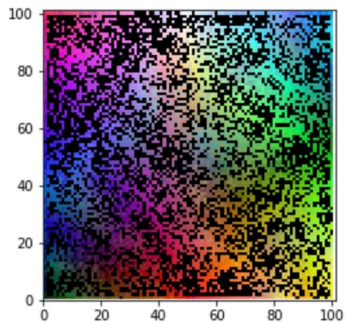

Fig 4. Classification of RGB colors using GEMA.

The second use case demonstrates that GEMA can be used to classify images. In this case, the map has been trained with the well-known dataset Modified National Institute of Standards and Technology (MNITS)[12]. This consists of images of 28x28 pixels with handwritten digits in black and white. In total, 60,000 examples are used for training and 10,000 as a test set. In Fig 5, it can be seen that a map with size 25 trained with 40,000 epochs is clustering well the different images. For example, images of number seven are clustered around the left upper corner. Another interesting point is that near a cluster of fours, there some nines that are quite difficult to distinguish. This occurs due to the morphological characteristics of both numbers.

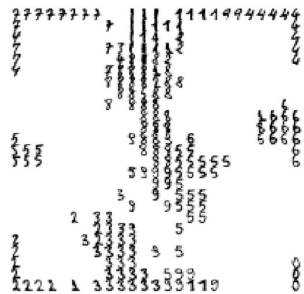

Fig 5. Handwritten numbers classified.

Another use case is the iris plants. This is also one of the most famous datasets in machine learning. It is a dataset introduced by Ronald Fisher in 1936, [12]. It consists of 50 samples of iris plants of three types measuring the width and length of sepals and petals. In the results provided in Fig 6, it can be seen that there are three different clusters, each corresponding to the different types of iris.

---

[12] http://yann.lecun.com/exdb/mnist

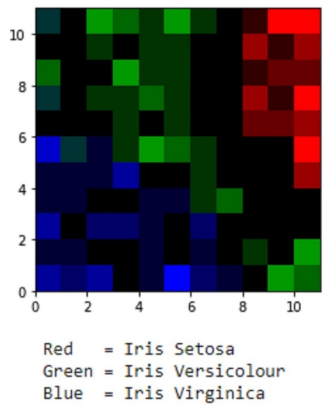

Fig 6. Iris flowers classified.

## 5. Conclusions and Outlook

The main aim was to develop a library to work with SOMs, also called Kohonen maps. This kind of neural model is divided into two main stages training and classification. Apart from developing them, other methods have been coded. It exists the possibility of obtaining a different kind of reports. The results can also be shown by plotting some interactive graphs. Finally, the map can be saved to use in the future by loading it.

In future works, the new implementation will be done. For example, other concepts defined by Kohonen like Neural Gas or Growing Gas. Also, the modules for obtaining reports and visualizing results will be extended. Finally, the library will be used in new real cases like clustering students depending on their psychological characteristics or possible food alerts depending on the characteristics of how they are transported around the European Union.

**References.**
1. Musumeci, F., Rottondi, C., Nag, A., Macaluso, I., Zibar, D., Ruffini, M., & Tornatore, M. (2018). An overview of application of machine learning techniques in optical networks. *IEEE Communications Surveys & Tutorials*, *21*(2), 1383-1408.
2. Kohonen T. Self-organized formation of topologically correct feature maps. Biological Cybernetics. 1982; 43(1):59-69. doi:10.1007/BF00337288
3. Kohonen, T. (1990). The self-organizing map. *Proceedings of the IEEE*, *78*(9), 1464-1480.
4. Cottrell, M., Olteanu, M., Rossi, F., & Villa-Vialaneix, N. (2018). Self-OrganizingMaps, theory and applications. *Investigación Operacional*, 2018, vol. 39, no 1, p. 1-23.
5. McKinney W (2011). pandas: a foundational Python library for data analysis and statistics. Python for High Performance and Scientific Computing, 1-9.
6. Hunter DJ (2007). Matplotlib: A 2D graphics environment. Computer Science Engineering. 2007; 9:90-95
7. Pedregosa, F., Varoquaux, G., Gramfort, A., Michel, V., Thirion, B., Grisel, O., & Vanderplas, J. (2011). Scikit-learn: Machine learning in Python. *the Journal of machine Learning research*, *12*, 2825-2830.
8. Lam, S. K., Pitrou, A., & Seibert, S. (2015, November).: A llvm-based python jit compiler. In *Proceedings of the Second Workshop on the LLVM Compiler Infrastructure in HPC* (pp. 1-6).
9. Wittek, P., Gao, S. C., Lim, I. S., & Zhao, L. (2013). Somoclu: An Efficient Parallel Library for Self-Organizing Maps. Journal of Statistical Software, 78: 1-21.
10. Hanke, M., Halchenko, Y. O., Sederberg, P. B., Hanson, S. J., Haxby, J. V., & Pollmann, S. (2009). PyMVPA: A Python toolbox for multivariate pattern analysis of fMRI data. Neuroinformatics, 7: 37-53.
11. Nogales A., García-Tejedor Á.J., Sanz N.M., de Dios Alija T. (2020) Competencies in Higher Education: A Feature Analysis with Self-Organizing Maps. In: Vellido A., Gibert K., Angulo C., Martín Guerrero J. (eds) Advances in Self-Organizing Maps, Learning Vector Quantization, Clustering and Data Visualization. WSOM 2019. Advances in Intelligent Systems and Computing, vol 976. Springer, Cham. https://doi.org/10.1007/978-3-030-19642-4_8
12. Fisher, R. A., & Marshall, M. (1936). Iris data set. *RA Fisher, UC Irvine Machine Learning Repository*, *440*, 87.